\documentclass{article}
\usepackage{spconf,amsmath,graphicx}

\usepackage{todonotes}
\usepackage[utf8]{inputenc} %
\usepackage[T1]{fontenc}
\usepackage{tabularx}

\usepackage{lineno,hyperref}
\usepackage{cite}
\usepackage{amsmath,graphicx}
\usepackage{float}
\usepackage{amssymb,amsmath,comment, graphicx,amsmath,mathrsfs, array, hyperref, multirow}

\usepackage{booktabs}
\usepackage{diagbox}
\usepackage{xcolor}
\usepackage[style=base]{caption}

\title{Joint prediction of truecasing and punctuation for conversational speech in low-resource scenarios}

\name{Raghavendra Pappagari$^1$, Piotr \.Zelasko$^{1}$, Agnieszka Mikołajczyk$^{2}$, Piotr P\k{e}zik$^{2}$, Najim Dehak$^{1}$}
\address{$^1$Center for Language and Speech Processing, Johns Hopkins University, Baltimore, USA \\
$^2$VoiceLab, Poland \\
\tt \{rpappag1, pzelasko,  ndehak3\}@jhu.edu \\
\tt \{agnieszka.mikolajczyk, piotr.pezik\}@voicelab.ai}

\begin{document}
\maketitle
\begin{abstract}

Capitalization and punctuation are important cues for comprehending written texts and conversational transcripts. 
Yet, many ASR systems do not produce punctuated and case-formatted speech transcripts.
We propose to use a multi-task system that can exploit the relations between casing and punctuation to improve their prediction performance. 
Whereas text data for predicting punctuation and truecasing is seemingly abundant, we argue that written text resources are inadequate as training data for conversational models.
We quantify the mismatch between written and conversational text domains by comparing the joint distributions of punctuation and word cases, and by testing our model cross-domain.
Further, we show that by training the model in the written text domain and then transfer learning to conversations, we can achieve reasonable performance with less data.

\end{abstract}

\section{Introduction}
\label{sec:intro}

Automatic speech recognition (ASR) systems are frequently used to transcribe meetings, calls, or dictated notes.
Most ASR models are trained to predict all lowercase or all uppercase transcripts without any punctuation.
Lack of text formatting makes it more difficult to read and comprehend text, even when it is free of speech recognition errors~\cite{grindlay2002missing}.
In addition to improving the legibility of speech transcripts, punctuation and capitalization restoration also facilitates other natural language processing (NLP) tasks, such as named entity recognition (NER)~\cite{lita2003truecasing}, part-of-speech tagging, syntactic parsing, and discourse segmentation.
Recently, \cite{zelasko2021whathelpstransformers} have shown that the presence of punctuation and truecasing is the single largest factor affecting the accuracy of various dialog act segmentation and classification systems. Furthermore, they found that punctuation is highly correlated with the presence of dialog act classes that correspond to conversational cues, such as incomplete utterances, restarts, repairs, and backchannels.

It is clear that the presence of punctuation is important for automated transcript processing -- but does there exist a single, correct way of inserting the punctuation symbols?
Truecasing and punctuation are abundantly available in written text resources. In some languages (such as Polish) the use of punctuation is well-defined by a set of rules. Other languages, such as English, permit -- to some extent -- arbitrary use of punctuation (e.g., the insertion of commas).
However, the use of truecasing and punctuation in speech transcripts is more complex due to the increased syntactic complexity of speech~\cite{kempson2000dynamic}.
Annotation of speech transcripts is a difficult and time-consuming task, which requires expertise and complex annotation guidelines.
We expect that the level of inter-annotator agreement for inserting punctuation is likely to be lower for conversational data, due to its spontaneity, disfluencies, and often unfinished sentences~\cite{kempson2000dynamic,charniak2001edit,purver-etal-2009-split}.
We hypothesize that there is a distribution mismatch between truecasing and punctuation present in written text resources, such as books, and in conversational transcripts.
In addition, conversational transcripts are typically much more difficult to come by -- except for abundantly-resourced languages such as English.

To investigate our hypothesis, we prepare an experimental setup with two text corpora that are representative of their respective domains -- the Gutenberg project books~\cite{gerlach2020standardized} and Fisher conversational transcripts~\cite{cieri2004fisher}. We consider a multi-task approach for joint punctuation and truecasing prediction to leverage the label correlations and dependencies between these tasks. 
We specifically investigate the following research questions:
\begin{itemize}
    \item How mismatched are book text and conversational transcripts domains in terms of punctuation and truecasing use?
    \item Can this mismatch be mitigated by leveraging cross-domain transfer learning, and to what extent?
    \item How much conversational data is needed to effectively leverage transfer learning from books? Note that we specifically choose the Fisher corpus due to its large size to answer this question.
    \item How much improvement can we expect from multi-task learning and is it consistent in both the books domain and the conversational domain?
    \item Does seeing truecased text in pretraining improve the performance of further fine-tuning for truecasing prediction?
\end{itemize}

The rest of the paper is organized as follows: in section~\ref{sec:related_work} we briefly highlight related literature; sections~\ref{sec:methods} and~\ref{sec:experiments} describe our methods and experimental setup. We present our results in section~\ref{sec:results} and conclude our findings in section~\ref{sec:conclusions}.

\section{Related work}
\label{sec:related_work}

The early efforts tackle truecasing and punctuation prediction using n-gram language models~\cite{lita2003truecasing,gravano2009restoring}.
However, the performance of simple n-gram language models suffers when long-range lexical information is required to disambiguate between punctuation classes~\cite{beeferman1998cyberpunc}.
Joint modelling of truecasing and punctuation tasks is considered in~\cite{sunkara2020robust,pahuja2017joint} using deep learning models in a classification framework.
Authors in~\cite{sunkara2020robust} assume punctuation as an independent task and truecasing as conditionally dependent on punctuation given latent representation of the input.
However, it is treated as a multi-task problem in~\cite{pahuja2017joint} where both truecasing and punctuation are independent given the input latent representation. 
Recently, \cite{o2021spgispeech} proposed training ASR models to directly transcribe text with truecasing and punctuation, which is enabled by the release of a large speech corpus with rich-format transcriptions. Unfortunately, their data use is not permissible for commercial applications.
Speech signal holds some cues such as pauses and intonation patterns to predict punctuation marks~\cite{levy2012effect}.
Incorporation of speech cues to the text-based models is explored in~\cite{zelasko2018punctuation, sunkara2020multimodal} and have shown improvements in punctuation prediction.
The distribution mismatch between text and conversational domains can be mitigated by retrofitting word embeddings to the target domain~\cite{augustyniak2020punctuation} when GloVe~\cite{pennington2014glove} embeddings are used in the model.

\section{Our approach}
\label{sec:methods}

Recent works have shown significant improvements by fine-tuning pre-trained language models for NLP-related downstream tasks~\cite{devlin2018bert, dai2019transformer, liu2019roberta}.
Inspired by this research, we choose to follow a similar method for truecasing and punctuation prediction.
In this work, we fine-tune base versions of BERT~\cite{devlin2018bert} pre-trained models for truecasing and punctuation prediction.
BERT base model consists of a sequence of 12 blocks of self-attention layer and fully-connected layer.
It is trained to optimize masked language model (MLM) objective and next sentence prediction (NSP).
For fine-tuning, we remove MLM and NSP heads from the model and only consider the encoder part for further modelling.
Below, we explain the procedure followed in this work to adapt the BERT model for joint prediction of truecasing and punctuation.

\subsection{Multi-tasking of truecasing and punctuation}

\begin{figure}
    \centering
    \includegraphics[width=0.5\textwidth]{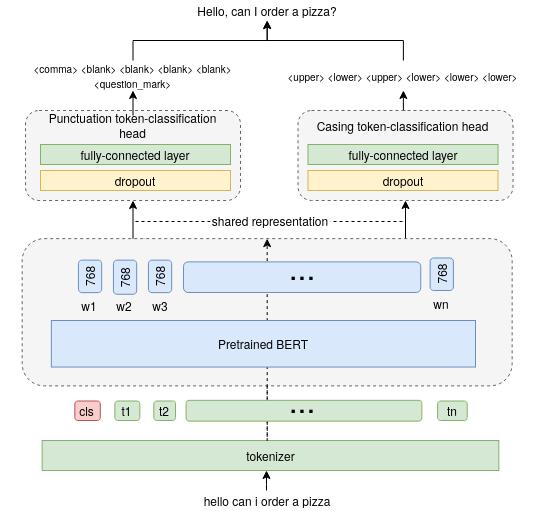}
    \caption{Flow chart for joint prediction of truecasing and punctuation}
    \label{fig:flowchart}
\end{figure}

In this work, we optimize truecasing and punctuation loss functions together in a multi-tasking manner to exploit relations between truecasing and punctuation.
Flow chart for the multi-task model with an example is shown in Fig.~\ref{fig:flowchart}.
As shown in the figure, we process the BERT encoder output using task-specific layers to obtain model predictions for our tasks.
Task-specific layers include a dropout layer and a fully connected layer. 
As the BERT encoder is common for both tasks, it has to encode information related to both the tasks and then task-specific layers retain relevant information depending on the task.
Through optimization of the loss computed for each task, we adapt the pre-trained weights and also learn the newly introduced parameters.
Loss for each task is computed by comparing fully connected layers output against the corresponding ground truth labels.
We use the cross-entropy loss function to compute losses as the targets for both the tasks are categorical.
For joint optimization, we minimize a weighted combination of both losses as shown in~\eqref{eq:multi_task_obj}.
We use a hyper-parameter $\lambda$ to balance/control the relative learning of the tasks. 
We show experiments with various values of $\lambda$.

\begin{equation}
    CE_{joint}=\lambda CE_{c} + (1-\lambda) CE_{p} \label{eq:multi_task_obj}
\end{equation}
where $CE_{c}$ and $CE_{p}$ are the cross-entropy loss functions for truecasing and punctuation respectively.

We use truecasing and punctuation models trained without multi-tasking as our baselines. 
We obtain baseline models for truecasing and punctuation by setting $\lambda$ to one and zero respectively.

To achieve truecasing and punctuation, we assign labels for every word in the input document and force the model to make a prediction for every word.
However, a word could be divided into several sub-word tokens as the BERT models use sub-word tokens at the input.
We compute loss only for the first sub-word token and do not consider the losses of other sub-word tokens.
We do not discard other sub-word tokens from training as they could be helpful to disambiguate the first sub-word token from other similar sub-word tokens.
In our experiments, we found that computing loss for the first or last sub-token token did not matter.
Loss for an input document is computed by taking a summation of each word loss which further used to calculate the batch loss.

The efficacy of fine-tuned models for downstream tasks could depend on the training data used for pre-training.
The training data used for the BERT base model include BooksCorpus~\cite{zhu2015aligning} and English Wikipedia.
Considering one of the target tasks in this paper, truecasing, it is interesting to see how the casing during pre-training affects the target tasks performance.
In this work, we compare BERT models trained with and without casing when adapted for truecasing and punctuation tasks.

\subsection{Exploration of model performance on conversational text in low-resource scenarios}

\begin{figure}
    \centering
    \includegraphics[width=0.45\textwidth]{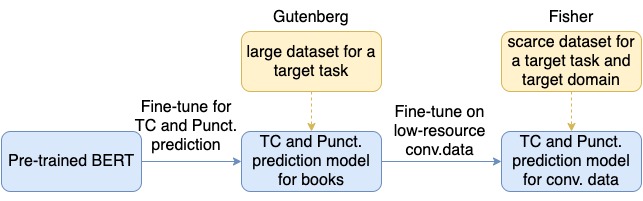}
    \caption{Fine-tuning process for low-resource scenarios. TC denotes truecasing; Punct. denotes punctuation; conv. denotes conversational}
    \label{fig:finetuning_lowresource}
\end{figure}

Given the effort required, minimizing the amount of annotated conversational data needed to achieve strong model performance is desirable in practical applications.
For that reason, we investigate the performance on various subsets of the Fisher dataset, most of which can be considered low resource scenarios.
Fig.~\ref{fig:finetuning_lowresource} shows the pipeline we follow in our experiments to achieve truecasing and punctuation prediction in low-resource scenarios.
We explore knowledge transfer from models trained on written text resources to conversational transcripts.
For this purpose, we use truecasing and punctuation prediction on Gutenberg dataset as an intermediate task.
That is, we first fine-tune the BERT-uncased model on the Gutenberg dataset to learn truecasing and punctuation patterns in the dataset then again fine-tune on small amounts of Fisher training data.
We compare it with a model obtained without any intermediate task i.e., fine-tuning BERT-uncased model directly on Fisher dataset.
We hypothesize that intermediate task optimization boosts performance on the Fisher dataset when only a few documents of Fisher are available. Ideally, a small number of conversations would be sufficient to bridge the gap between the two domains.

\begin{table}
\centering
\caption{Label count for each class in our datasets. Perc. denotes percentage of the label}
\label{tab:label_count_dataset}
\resizebox{\columnwidth}{!}{%
\begin{tabular}{@{}l|l|ll|ll@{}}
\midrule
 &    &      Fisher &    &      \multicolumn{2}{c}{Gutenberg} \\ 
\midrule
 &             & Count    & Perc. (\%)  & Count    & Perc. (\%) \\
\midrule
\multirow{8}{*}{\begin{tabular}[c]{@{}c@{}}Punctuation\\ marks\end{tabular}} &
  Blank &
  12 425 398 &
  72.05 &
  11 019 746 &
  86.44 \\
 & Comma       & 2 047 050  & 11.87   & 955 422   & 7.49  \\
 & Ellipsis    & 44 077    & 0.26   & 2 928     & 0.02  \\
 & Exclamation & 9 922     & 0.06   & 31 731    & 0.25  \\
 & FullStop    & 1 454 839  & 8.44    & 559 424   & 4.39  \\
 & Question    & 220 998   & 1.28   & 32 677    & 0.26  \\
 & SemiColon   & 2 214     & 0.01   & 89 262    & 0.7   \\
 & DoubleDash  & 1 041 119  & 6.04    & 57 198    & 0.45  \\
 \midrule
\multirow{3}{*}{\begin{tabular}[c]{@{}c@{}}Truecasing\\ classes\end{tabular}} &
  AUC &
  841 697 &
  4.88 &
  184 877 &
  1.45 \\
 & LC          & 14 259 782 & 82.69  & 11 400 740 & 89.43 \\
 & UC          & 2 144 138  & 12.43   & 1 162 771  & 9.12  \\ 
 \bottomrule
\end{tabular}%
}
\end{table}

\section{Experimental setup}
\label{sec:experiments}

\subsection{Corpora}
\label{sec:experiments:corpora}

\begin{figure}
    \centering
    \includegraphics[width=0.45\textwidth]{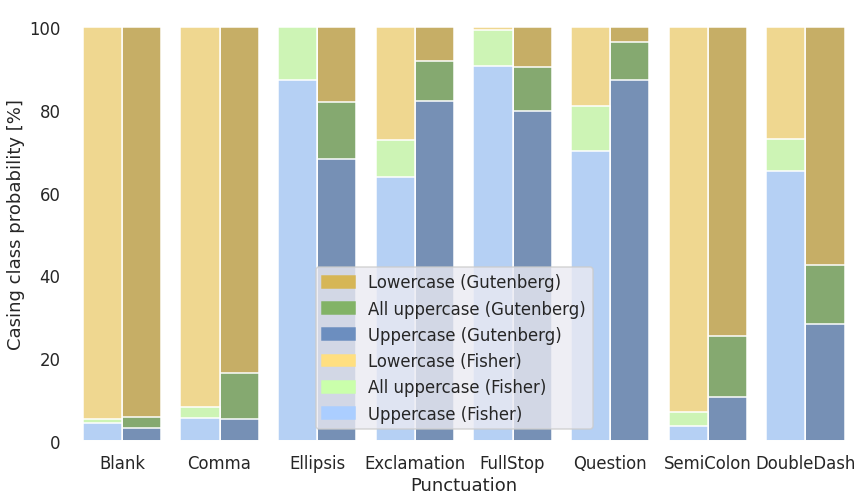}
    \caption{Distribution of truecasing classes after a punctuation occurs in Fisher and Gutenberg datasets. For example, more than 90\% of the words after Blank have LowerCase; around 60\% of the words after Exclamtaion have UpperCase in Fisher.}
    \label{fig:datasets_punct_vs_TC}
\end{figure}

\textbf{The Fisher corpus}~\cite{cieri2004fisher} includes telephone call recordings along with rich manual transcriptions including capitalization and punctuation. 
Participants across a variety of demographic categories had conversations about randomly pre-assigned topics to maximize inter-speaker variation and vocabulary breadth. 
Transcriptions along with capitalization and punctuation marks were manually introduced by annotators based on automatically segmented recordings, where punctuation marks denote the pauses and intonations. 
The Fisher dataset consists of 9168 transcripts for training, 509 for development and 510 for testing.

\textbf{The Gutenberg corpus} is a collection of over 50k fictional, multilingual books~\cite{gerlach2020standardized}. 
It contains copyright-free books including biographies, fantasy, art, fiction, poetry and philosophy. The dataset has rich and high-quality punctuation annotation that was double-checked by the editor's team. 
However, due to the license restrictions, most of the books are dated between 1800-1930 making the language possibly outdated, especially when considering truecasing prediction.
In this work, we use 75\%, 5\% and 25\% splits for model training, development and testing.

\subsection{Data preparation for truecasing and punctuation}

We assign one truecasing and one punctuation label for every word.
For truecasing, we consider three labels, namely all upper casing (AUC), lower casing (LC) and upper casing (UC). 
AUC is assigned for words with all upper case characters; LC is assigned for words with all lower case characters; UC is assigned for words starting with an upper case character and also for words with mixed casing.
For punctuation, we assign a punctuation label out of 8 possible labels for each word.
The list of punctuation considered in this work are Blank, Full-stop (.), Comma (,), Question mark (?), Exclamation mark (!), Semicolon (;), Double-dash (--), Ellipsis (...).
In cases where punctuation appears after a word, we assign that word with that punctuation otherwise we assign Blank. 
Note that this setup diverges from previous works on the Fisher corpus~\cite{zelasko2018punctuation,augustyniak2020punctuation,sunkara2020multimodal} -- we included more punctuation classes specifically to investigate how well the models deal with less frequent punctuation marks that exist in both domains.

Class count and their percentage in the corresponding dataset are shown in Table~\ref{tab:label_count_dataset}.
The distribution of the token count for punctuation and truecasing classes are highly skewed.
As expected, LC class frequency is much higher than AUC and UC in both datasets.
Truecasing classes distribution between the datasets is more similar compared to punctuation classes distribution.
Blank dominates in both the datasets followed by Comma and FullStop.
One noticeable difference between the datasets is the use of DoubleDash: it occurs for 6.04\% of the tokens in Fisher whereas only for 0.45\% of the tokens in the Gutenberg dataset.
We observed that DoubleDash is majorly used at the change of turns in the Fisher conversations.
Question mark usage is more frequent in Fisher compared to the Gutenberg dataset which could be expected -- questions are the main tool of eliciting information from the other parties in conversations.
Ellipsis, although rare, is significantly more frequent in conversations, reflecting a larger portion of unfinished utterances.
Finally, semicolons are practically non-existent in conversation transcripts.

\section{Results}
\label{sec:results}

In this section, we present the results of our experiments on truecasing and punctuation. 
First, we show the impact of casing and punctuation in the input document on each other's task followed by a multi-task model to improve the performance of both tasks.
Then, we explore the model performance on conversational text when only limited data is available.

\subsection{Correlation between truecasing and punctuation}
\label{subsec:correlation_truecasing_punctuation}

\begin{table}
\caption{Effect of punctuation and truecasing on each other's prediction. We adapt BERT-cased model for this study. All numbers denote Macro F1 scores.} \label{table:effect_punct_casing}
\resizebox{\columnwidth}{!}{
\begin{tabular}{c|c|c|c|c}
\toprule
\multirow{2}{*}{Task}        & \multicolumn{2}{c|}{Input documents with} & \multirow{2}{*}{Gutenberg } & \multirow{2}{*}{Fisher} \\
                             & Punctuation       & Casing      &          &                       \\
                             \midrule
\multirow{2}{*}{Truecasing}  & Yes               & No          &  97.23        &      97.54                \\
                             & No                & No          &    94.33      & 92.67                    \\
                             \midrule
\multirow{2}{*}{Punctuation} & No                & Yes         &    80.16      & 50.49                    \\
                             & No                & No          & 75.10         &  47.61                \\
                             \bottomrule
\end{tabular}
}
\end{table}

Fig.~\ref{fig:datasets_punct_vs_TC} presents the statistics of word casing for the next word after punctuation for Fisher and Gutenberg datasets.
As is the case with some of the basic rules in the English language, we can observe that the most frequent casing after the full stop, question mark and exclamation is the upper casing.
Similarly, lowercased words follow the majority of the time after comma, semicolon and Blank.
We can observe that the most frequent casing after the double dash is different in both datasets: a lower casing (57.41\%) in Fisher and an upper casing (65.38\%) in Gutenberg.
Calculating Macro F1 with the most frequent casing, we see 54.85\% for Fisher and 52.7\% for the Gutenberg dataset. 
In this case, we almost never predict the AUC class and miss many UC words which are likely the most important classes for applications like named entity recognition.

Table~\ref{table:effect_punct_casing} shows the impact of truecasing and punctuation on each other's prediction.
To evaluate the effect of punctuation on truecasing prediction, we experiment with and without including punctuation in the input documents.
Similarly, to quantify the effect of casing on punctuation prediction, we experiment with and without including casing in the input documents.
We fine-tune the BERT-cased model for this experiment as the tokenization process removes word casing in the BERT-uncased model.
We can observe that the truecasing task performance is dropped by 2.9\% and 4.87\% absolute on Gutenberg and Fisher respectively without punctuation in the input.
Punctuation task performance is dropped by 5.06\% and 2.88\% absolute on Gutenberg and Fisher respectively without casing in the input.
The performance drop can be explained by the data statistics presented in Fig.~\ref{fig:datasets_punct_vs_TC}.
Results of this experiment along with Fig.~\ref{fig:datasets_punct_vs_TC} strongly suggest that truecasing and punctuation are correlated.
Hence, we expect their joint modelling improves the performance of both tasks.

\begin{table}
\centering
\caption{Results with multi-tasking using \textbf{BERT-uncased} pre-trained model for fine-tuning. TC -- Truecasing task; Punct. -- Punctuation task. Best results (Macro F1 scores) in each column are in bold.}
\label{tab:multi_tasking_BERT_uncased}
\resizebox{0.8\columnwidth}{!}{%
\begin{tabular}{@{}c||c|c||c|c@{}}
\toprule
    & \multicolumn{2}{c||}{Fisher} &  \multicolumn{2}{c}{Gutenberg } \\ 
    \cmidrule{2-5} 
Lambda & TC             & Punct.               & TC     & Punct.          \\
\midrule
1      & 92.92          & -                      & 95.45  & -               \\
0.9 & \textbf{93.06}   & 46.86          & \textbf{95.85}       & 73.45       \\
0.75   & \textbf{93.06} & 47.56           & 95.80  & 75.64           \\
0.5    & 93.02          & \textbf{48.06}      & 95.71  & 76.23           \\
0.25   & 92.89          & 47.70                  & 95.51  & 76.61           \\
0.1    & 92.66         & 47.65        & 95.24 & 76.76          \\
0      & -              & 47.08        & -      & \textbf{77.58} \\
\bottomrule
\end{tabular}%
}
\end{table}

\subsection{Experiments with multi-tasking}

Tables~\ref{tab:multi_tasking_BERT_uncased} and~\ref{tab:multi_tasking_BERT_cased} present the results of multi-task models when fine-tuned from BERT-uncased and BERT-cased pre-trained models respectively along with corresponding baselines.
The multi-task model is trained with the objective function shown in~\eqref{eq:multi_task_obj} where setting $\lambda$ to 0 provides a baseline for truecasing and 1 for punctuation task.
We show experiments with $\lambda=0.1, 0.25, 0.5, 0.75, 0.9$ for multi-task model to find the optimum values for $\lambda$.
For truecasing task, multi-tasking provides improvements with $\lambda=0.9, 0.75, 0.5$ and sometimes even when $\lambda$ is less than 0.5.
We can observe improved performance for punctuation task too with $\lambda=0.1$ in all cases except when BERT-uncased is fine-tuned for Gutenberg.
It supports our hypothesis that joint modelling helps both truecasing and punctuation tasks.

Comparing BERT-cased (Table~\ref{tab:multi_tasking_BERT_cased}) and BERT-uncased (Table~\ref{tab:multi_tasking_BERT_uncased}) pre-trained models, the latter suited better for truecasing task.
The BERT-cased model was trained on the cased text and the BERT-uncased model on uncased text.
For the punctuation task, the BERT-uncased model provided better results on both Gutenberg and Fisher compared to BERT-cased model except for the Fisher baseline.
From these results, we can conclude that pre-training with uncased text is more suitable for both truecasing and punctuation tasks compared to training with cased text.

We can observe that punctuation task performance ranges are quite different for Fisher (around 47\%) and Gutenberg dataset (around 75\%).
We suspect two reasons for the difference: 1) Gutenberg is a collection of books, usually well proof-read, we can expect consistent punctuation compared to the punctuation in conversations and 2) BERT models are pre-trained on written text.
Note that it is a mere comparison demonstrating the difficulty of recognizing punctuation in written and conversational text but not to be taken in a strict sense as the corresponding models are trained and tested on different data.

\begin{table}
\centering
\caption{Results with multi-tasking using \textbf{BERT-cased} pre-trained model for fine-tuning. TC -- Truecasing task; Punct. -- Punctuation task. Best results (Macro F1 scores) in each column are in bold.}
\label{tab:multi_tasking_BERT_cased}
\resizebox{0.8\columnwidth}{!}{%
\begin{tabular}{@{}c||c|c||c|c@{}}
\toprule
    & \multicolumn{2}{c||}{Fisher}  & \multicolumn{2}{c}{Gutenberg } \\
    \cmidrule{2-5} 
Lambda & TC             & Punct.           & TC     & Punct.           \\
\midrule
1    & 92.67          & -              & 94.33          & -              \\
0.9  & \textbf{93.00} & 46.56        & \textbf{95.47} & 71.93          \\
0.75 & 92.96          & 47.37             & 95.39          & 72.84          \\
0.5  & 92.93          & 47.58             & 95.21          & 75.40          \\
0.25 & 92.77          & 47.49         & 94.98          & 75.09          \\
0.1  & 92.51          & \textbf{47.63}          & 94.85          & \textbf{75.86} \\
0    & -              & 47.61                   & \textbf{-}     & 75.10          \\
\bottomrule
\end{tabular}
}
\end{table}

\subsection{Experiments on Fisher in low-resource scenarios}

\begin{table}
\centering
\caption{Results on Fisher test set with various training dataset sizes. With Interm. Task denotes fine-tuning is on intermediate task model; W/O Interm. Task denotes fine-tuning on BERT-uncased model directly without intermediate task. * denotes the corresponding model is evaluated without fine-tuning.}
\label{tab:fisher_lowresource_setting}
\resizebox{\columnwidth}{!}{%
\begin{tabular}{@{}c|c|c|c|c@{}}
\toprule
\multirow{2}{*}{\begin{tabular}[c]{@{}c@{}}\#Fisher \\ documents\end{tabular}} & \multicolumn{2}{c|}{Truecasing}   & \multicolumn{2}{c}{Punctuation}  \\
    \cmidrule{2-5} 

                                   & With Interm. Task  & W/O Interm. Task  & With Interm. Task & W/O Interm. Task  \\
                                   \midrule
0       & \, \, 73.95* & -     & \, \, 21.62* & -     \\
50      & 72.76  & 32.17 & 26.22  & 9.74  \\
100     & 88.69  & 76.33 & 37.93  & 22.65 \\
250     & 90.98  & 89.29 & 42.39  & 35.09 \\
500  & 91.68  & 91.07 & 43.64  & 42.91 \\
1000 & 92.08  & 91.76 & 45.01  & 45.13 \\
5000 & 92.76  & 92.78 & 47.4   & 46.81 \\
9168 & \textbf{92.99}  & \textbf{93.02} & \textbf{47.83}  & \textbf{48.06} \\
\bottomrule
\end{tabular}%
}
\end{table}

\begin{table*}[ht]
\centering
\caption{Comparison of punctuation class-wise f1-scores of models fine-tuned with different amounts of Fisher data.}
\label{tab:classwise_f1_lowresource}
\resizebox{\textwidth}{!}{%
\begin{tabular}{@{}c|cccccccc|cccccccc@{}}
\toprule
\multirow{2}{*}{\begin{tabular}[c]{@{}c@{}}\#Fisher \\ Documents\end{tabular}}& \multicolumn{8}{c|}{With Intermediate Task} & \multicolumn{8}{c}{Without Intermediate Task} \\
 \cmidrule{2-17}
 & Blank & Comma & Ellipsis & Exclamation & FullStop & Question & SemiColon & DoubleDash & Blank & Comma & Ellipsis & Exclamation & FullStop & Question & SemiColon & DoubleDash \\
  \midrule
  0 &   91.81 & 46.93 &  0 &   0  &  7.01 &   27.21 &  0 &   0 & - & - & - & - & - & - &  - & - \\
 50   & 91.48 & 42.92 & 0  & 1.66 & 42.93 & 30.79 & 0 & 0.01   & 53.37 & 7.71  & 0.31  & 0.07 & 9.63  & 0     & 0    & 6.80   \\
100  & 93.65 & 55.67 & 0     & 0    & 63.50  & 49.69 & 0    & 40.94  & 91.74 & 44.49 & 0  & 0 & 44.97 & 0  & 0 & 0.02  \\
250  & 94.50  & 60.55 & 0     & 0    & 67.53 & 58.00    & 0    & 58.55 & 93.91 & 57.05 & 0     & 0    & 62.43 & 12.12 & 0    & 55.23 \\
500  & 94.68 & 61.78 & 0     & 0.63 & 68.91 & 61.61 & 0    & 61.49  & 94.42 & 59.91 & 6.14  & 0    & 66.97 & 55.72 & 0    & 60.10  \\
1000 & 94.87 & 62.89 & 3.98  & 0.64 & 69.94 & 63.96 & 0    & 63.81 & 94.73 & 62.15 & 10.28 & 0    & 69.01 & 61.90  & 0    & 62.97 \\
5000 & 95.17 & 64.56 & 12.23 & 0.65 & 71.72 & 67.67 & 0    & 67.24  & 95.12 & 64.21 & 7.49  & 0.70  & 71.65 & 67.75 & 0    & 67.56 \\
9168 & 95.25 & 64.88 & 10.83 & 3.14 & 72.22 & 68.61 & 0    & 67.74 & 95.27 & 65.06 & 10.58 & 3.28 & 72.4  & 69.34 & 0    & 68.53 \\
    \bottomrule
\end{tabular}%
}
\end{table*}

We evaluate the re-usability of truecasing and punctuation models trained on written text (Gutenberg) for conversational text (Fisher). 
For this purpose, we fine-tune the BERT-uncased model on the Gutenberg dataset with $\lambda$ equal to 0.5 as an intermediate task (refer Fig.~\ref{fig:finetuning_lowresource}).
Then, we fine-tune again on a limited number of transcriptions from Fisher and compare it with the BERT-uncased model fine-tuned on Fisher directly.
Table~\ref{tab:fisher_lowresource_setting} shows the results on the Fisher dataset when only a limited number of annotated conversational documents are available.
Each row, for example 2$^{nd}$ row can be interpreted as -- we achieve 72.76\% truecasing performance on Fisher test set when using an intermediate task and only 50 Fisher documents are available for training.
Similarly, 32.17\% in the same row can be read as -- we achieve 32.17\% truecasing performance when we fine-tune the BERT-uncased model on only 50 Fisher documents directly without any intermediate task.
For both tasks, the performance improved with more training data size.
We can observe significant performance improvements when trained with 100 documents compared to training with 50 documents.
Adding more documents to the training after 250 documents has little effect on truecasing task.
Whereas for punctuation, the performance steadily improved with more training samples although relatively less improvement after 500 documents.

Table~\ref{tab:classwise_f1_lowresource} shows class-wise f1-scores for punctuation corresponding to the Macro F1 scores presented in Table~\ref{tab:fisher_lowresource_setting}.
We can observe 91.81\%, 46.93\% and 27.21\% for Blank, Comma and Question mark when the intermediate task model (trained with Gutenberg ) is evaluated on Fisher documents.
Significant improvements to the punctuation marks FullStop, Question and DoubleDash have come at lower training data sizes (50-250 documents) and then a steady improvement with more training documents.
We can observe that fine-tuning the Gutenberg dataset model provided better results compared to fine-tuning the BERT-uncased model in most cases.
For Question mark, fine-tuning from the BERT-uncased model yielded only 12.12\% f1-score even with 250 Fisher documents whereas fine-tuning a written text model provided 30.79\% with just 50 documents suggesting that the usage of the Question mark is similar in both Gutenberg and Fisher datasets.
A recall of 45.32\% for FullStop (f1-score of 42.93\% in Table~\ref{tab:classwise_f1_lowresource}) implies that reasonable sentence segmentation can be achieved with 50 spoken documents for training.

The most frequently used punctuation marks Blank, Comma, FullStop and Question are recognized decently with just 50 Fisher documents when the Gutenberg dataset model is fine-tuned.
From a practical point of view, it implies that most of the spoken documents can be enriched with punctuation even with very little annotated conversational data.
Other punctuation marks, Ellipsis, Exclamation and SemiColon are not recognized even with the larger set of training documents.
We suspect their inconsistent usage and less frequency in training data could have contributed to their poor performance.

\section{Conclusions and future work}
\label{sec:conclusions}

In this paper, we presented a multi-tasking model for truecasing and punctuation to take advantage of the correlations between them.
Our experiments have shown that multi-task modelling improves the performance of both tasks.
Experiments with fine-tuning BERT-based pre-trained models revealed that pre-training with casing provides inferior results compared to pre-training with uncased text on truecasing and punctuation prediction.
Through knowledge transfer from written text to conversational text models, we found that as little as 50 annotated spoken documents can provide decent performance for most frequently used punctuation marks such as Comma, FullStop, and Question mark.
And, adding more data provided larger improvements up until training set size 500 documents and smaller improvements after that.

In future work, we plan to investigate the transfer of knowledge from other datasets that are similar to conversational text to improve truecasing and punctuation task performance on conversational text.
We also look to explore cross-lingual knowledge transfer to minimize annotation costs in low-resource languages.

\bibliographystyle{IEEEbib}
\bibliography{main}

\begin{thebibliography}{10}

\bibitem{grindlay2002missing}
Benjamin James~William Grindlay,
\newblock {\em Missing the point: the effect of punctuation on reading
  performance/Benjamin JW Grindlay.},
\newblock Ph.D. thesis, Adelaide University, Adelaide, Australia, 2002.

\bibitem{lita2003truecasing}
Lucian~Vlad Lita, Abe Ittycheriah, Salim Roukos, and Nanda Kambhatla,
\newblock ``Truecasing,''
\newblock in {\em Proceedings of the 41st Annual Meeting of the Association for
  Computational Linguistics}, 2003, pp. 152--159.

\bibitem{zelasko2021whathelpstransformers}
Piotr Żelasko, Raghavendra Pappagari, and Najim Dehak,
\newblock ``What helps transformers recognize conversational structure?
  {I}mportance of context, punctuation, and labels in dialog act recognition,''
\newblock {\em Transactions of the Association of Computational Linguistics
  (accepted, to appear)}, 2021.

\bibitem{kempson2000dynamic}
Ruth Kempson, Wilfried Meyer-Viol, and Dov~M Gabbay,
\newblock {\em Dynamic syntax: The flow of language understanding},
\newblock Wiley-Blackwell, 2000.

\bibitem{charniak2001edit}
Eugene Charniak and Mark Johnson,
\newblock ``Edit detection and parsing for transcribed speech,''
\newblock in {\em Second Meeting of the North American Chapter of the
  Association for Computational Linguistics}, 2001.

\bibitem{purver-etal-2009-split}
Matthew Purver, Christine Howes, Eleni Gregoromichelaki, and Patrick Healey,
\newblock ``Split utterances in dialogue: a corpus study,''
\newblock in {\em Proceedings of the {SIGDIAL} 2009 Conference}, London, UK,
  Sept. 2009, pp. 262--271, Association for Computational Linguistics.

\bibitem{gerlach2020standardized}
Martin Gerlach and Francesc Font-Clos,
\newblock ``A standardized project gutenberg corpus for statistical analysis of
  natural language and quantitative linguistics,''
\newblock {\em Entropy}, vol. 22, no. 1, pp. 126, 2020.

\bibitem{cieri2004fisher}
Christopher Cieri, David Miller, and Kevin Walker,
\newblock ``The fisher corpus: A resource for the next generations of
  speech-to-text.,''
\newblock in {\em LREC}, 2004, vol.~4, pp. 69--71.

\bibitem{gravano2009restoring}
Agustin Gravano, Martin Jansche, and Michiel Bacchiani,
\newblock ``Restoring punctuation and capitalization in transcribed speech,''
\newblock in {\em 2009 IEEE International Conference on Acoustics, Speech and
  Signal Processing}. IEEE, 2009, pp. 4741--4744.

\bibitem{beeferman1998cyberpunc}
Doug Beeferman, Adam Berger, and John Lafferty,
\newblock ``Cyberpunc: A lightweight punctuation annotation system for
  speech,''
\newblock in {\em Proceedings of the 1998 IEEE International Conference on
  Acoustics, Speech and Signal Processing, ICASSP'98 (Cat. No. 98CH36181)}.
  IEEE, 1998, vol.~2, pp. 689--692.

\bibitem{sunkara2020robust}
Monica Sunkara, Srikanth Ronanki, Kalpit Dixit, Sravan Bodapati, and Katrin
  Kirchhoff,
\newblock ``Robust prediction of punctuation and truecasing for medical asr,''
\newblock in {\em Proceedings of the First Workshop on Natural Language
  Processing for Medical Conversations}, 2020, pp. 53--62.

\bibitem{pahuja2017joint}
Vardaan Pahuja, Anirban Laha, Shachar Mirkin, Vikas Raykar, Lili Kotlerman, and
  Guy Lev,
\newblock ``Joint learning of correlated sequence labeling tasks using
  bidirectional recurrent neural networks,''
\newblock {\em Proc. Interspeech 2017}, pp. 548--552, 2017.

\bibitem{o2021spgispeech}
Patrick~K O'Neill, Vitaly Lavrukhin, Somshubra Majumdar, Vahid Noroozi, Yuekai
  Zhang, Oleksii Kuchaiev, Jagadeesh Balam, Yuliya Dovzhenko, Keenan Freyberg,
  Michael~D Shulman, et~al.,
\newblock ``Spgispeech: 5,000 hours of transcribed financial audio for fully
  formatted end-to-end speech recognition,''
\newblock {\em arXiv preprint arXiv:2104.02014 (accepted at Interspeech 2021)},
  2021.

\bibitem{levy2012effect}
Tal Levy, Vered Silber-Varod, and Ami Moyal,
\newblock ``The effect of pitch, intensity and pause duration in punctuation
  detection,''
\newblock in {\em 2012 IEEE 27th Convention of Electrical and Electronics
  Engineers in Israel}. IEEE, 2012, pp. 1--4.

\bibitem{zelasko2018punctuation}
Piotr {\.Z}elasko, Piotr Szyma{\'n}ski, Jan Mizgajski, Adrian Szymczak, Yishay
  Carmiel, and Najim Dehak,
\newblock ``Punctuation prediction model for conversational speech,''
\newblock {\em Proc. Interspeech 2018}, pp. 2633--2637, 2018.

\bibitem{sunkara2020multimodal}
Monica Sunkara, Srikanth Ronanki, Dhanush Bekal, Sravan Bodapati, and Katrin
  Kirchhoff,
\newblock ``Multimodal semi-supervised learning framework for punctuation
  prediction in conversational speech,''
\newblock {\em Proc. Interspeech 2020}, pp. 4911--4915, 2020.

\bibitem{augustyniak2020punctuation}
{\L}ukasz Augustyniak, Piotr Szyma{\'n}ski, Miko{\l}aj Morzy, Piotr
  {\.Z}elasko, Adrian Szymczak, Jan Mizgajski, Yishay Carmiel, and Najim Dehak,
\newblock ``Punctuation prediction in spontaneous conversations: Can we
  mitigate asr errors with retrofitted word embeddings?,''
\newblock {\em Proc. Interspeech 2020}, pp. 4906--4910, 2020.

\bibitem{pennington2014glove}
Jeffrey Pennington, Richard Socher, and Christopher~D Manning,
\newblock ``Glove: Global vectors for word representation,''
\newblock in {\em Proceedings of the 2014 conference on empirical methods in
  natural language processing (EMNLP)}, 2014, pp. 1532--1543.

\bibitem{devlin2018bert}
Jacob Devlin, Ming-Wei Chang, Kenton Lee, and Kristina Toutanova,
\newblock ``Bert: Pre-training of deep bidirectional transformers for language
  understanding,''
\newblock {\em arXiv preprint arXiv:1810.04805}, 2018.

\bibitem{dai2019transformer}
Zihang Dai, Zhilin Yang, Yiming Yang, Jaime Carbonell, Quoc~V Le, and Ruslan
  Salakhutdinov,
\newblock ``Transformer-xl: Attentive language models beyond a fixed-length
  context,''
\newblock {\em arXiv preprint arXiv:1901.02860}, 2019.

\bibitem{liu2019roberta}
Yinhan Liu, Myle Ott, Naman Goyal, Jingfei Du, Mandar Joshi, Danqi Chen, Omer
  Levy, Mike Lewis, Luke Zettlemoyer, and Veselin Stoyanov,
\newblock ``Roberta: A robustly optimized bert pretraining approach,''
\newblock {\em arXiv preprint arXiv:1907.11692}, 2019.

\bibitem{zhu2015aligning}
Yukun Zhu, Ryan Kiros, Rich Zemel, Ruslan Salakhutdinov, Raquel Urtasun,
  Antonio Torralba, and Sanja Fidler,
\newblock ``Aligning books and movies: Towards story-like visual explanations
  by watching movies and reading books,''
\newblock in {\em Proceedings of the IEEE international conference on computer
  vision}, 2015, pp. 19--27.

\end{thebibliography}

\end{document}